\documentclass{llncs}

\usepackage{a4,a4wide}

\usepackage{graphicx}
\usepackage{amssymb}                      %
\usepackage{amsmath}                      %
\usepackage{array,booktabs,multirow}      
\usepackage{esvect}
\usepackage{color}                        %

\newcommand{\textaa}[1]{{\hbox{\texttt{#1}}}}    
\newcommand{\textmeta}[1]{{\ensuremath{\langle #1 \rangle}}}    

\newcommand{\WEG}[1]{}
\newcommand{\ol}[1]{\overline{#1}}

\renewcommand{\leq}{\leqslant}

\renewcommand{\geq}{\geqslant}

\newcommand{\cspR}{\ensuremath{\mathit{CR}(\R)}}
\newcommand{\cspOf}[1]{\ensuremath{\mathit{CR}(#1)}}
\newcommand{\solutionsR}{\ensuremath{\mathit{Sol}_\mathit{CR}(\R)}}
\newcommand{\solutionsOf}[1]{\ensuremath{\mathit{Sol}_\mathit{CR}(#1)}}
\newcommand{\ordnungInduzierteOCF}{\ensuremath{\preccurlyeq}}

\newcommand{\cspRprog}{\texttt{GenOCF}}
\newcommand{\induzierteOCF}[1]{\ensuremath{\kappa_{#1}}}
\newcommand{\syntheticKB}[1]{\texttt{kb\_synth<\(n\)>\_c<\(2n\!\!-\!\!1\)>.pl}}
\newcommand{\syntheticKBsymbol}[2]{\texttt{kb\_synth<\(\ensuremath{#1}\)>\_c<\ensuremath{#2}>.pl}}
\newcommand{\syntheticKBzahl}[2]{\texttt{kb\_synth{#1}\_c{#2}.pl}}

\newcommand{\cL}{\ensuremath{\mathcal L}}
\newcommand{\R}{\ensuremath{\mathcal R}}

\newcommand{\condAB}{(B | A)}

\newcommand{\notA}{\overline{A}}

\newcommand{\notB}{\overline{B}}
\newcommand{\condL}{\mbox{$(\cL \mid \cL)$}}

\newcommand{\rationals}{\mathbb{Q}}

\newcommand{\kappaiplus}[1]{\kappa_{#1}^+}
\newcommand{\kappaiminus}[1]{\kappa_{#1}^-}
\newcommand{\kappaiminusStrich}[1]{{\kappa'}_{#1}^{-}}
\newcommand{\kappaiminusVektor}{\ensuremath{\vv{\kappa}}}
\newcommand{\kappaiminusVektorStrich}{\ensuremath{\vv{\kappa}'}}

\newcommand{\beweisendezeichen}%
{\penalty50\hspace*{0pt plus 1fil}\parfillskip=0pt\mbox{$\Box$}}

\newcommand{\fussnoteOhneMarkierung}[1]%
{%
\footnote{#1}%
\addtocounter{footnote}{-1}%
}

\newcommand{\nichtMod}{\ensuremath{\,\,\,/\hspace{-1em}}}

\newcommand{\satzCL}[2]{\ensuremath{(#1|#2)}}

\newcommand{\conditional}[3]{\ensuremath{(#1|#2)[#3]}}
\newcommand{\satzPCL}[3]{\ensuremath{\conditional{#1}{#2}{#3}}}

\newlength{\abstand}
\newcommand{\Nat}{\ensuremath{\mathbb{N}}}

\newcommand{\displayFctEins}[3]{\ensuremath#1:\, #2 \rightarrow #3}

\newcommand{\universe}[1]{\textit{#1}}

\newcommand{\Vocabulary}{\ensuremath{\Sigma}}
\newcommand{\World}{\ensuremath{\Omega}}

\newcommand{\MeasureOCF}{\universe{\Nat}}
\newcommand{\symbolOCF}{\cal O}

\newcommand{\modelsQnoIocf}{\ensuremath{\, {\models}_{{\symbolOCF}} \, }}

\newcommand{\StateSymOCF}{\ensuremath{\kappa}}

\usepackage{fancyvrb}
\RecustomVerbatimEnvironment{Verbatim}{Verbatim}{xleftmargin=5mm}
\newenvironment{qverbatim}
{\quote\Verbatim}
{\endVerbatim\endquote}

\title{A Constraint Logic Programming Approach for Computing 
Ordinal Conditional Functions}
\author{Christoph Beierle$^{1}$, \ Gabriele Kern-Isberner$^{2}$, Karl S\"{o}dler$^{1}$}
\institute{\(^1\)Dept. of Computer Science, 
FernUniversit\"{a}t in Hagen,
58084 Hagen, Germany
\\
\(^2\)Dept. of Computer Science, TU Dortmund,
44221 Dortmund, Germany
}
\begin{document}
\maketitle

\begin{abstract}
In order to give appropriate semantics to qualitative conditionals
of the form \emph{if A then normally B}, ordinal conditional functions
(OCFs) ranking the possible worlds according to their degree
of plausibility can be used. 
An OCF accepting all conditionals of a knowledge base R can be
characterized as the solution of a constraint satisfaction problem.
We present a high-level, declarative approach using 
constraint logic programming techniques for solving this constraint 
satisfaction problem. In particular, the approach developed here supports 
the generation of all minimal solutions;
these minimal solutions are of special interest as they provide a 
basis for model-based inference from R.
\end{abstract}

\section{Introduction}
\label{sec_introduction}

In knowledge representation, rules play a prominent role. Default rules
of the form \emph{If A then normally B} are being investigated in
nonmonotonic reasoning, and various semantical approaches have been
proposed for such rules. Since it is not possible to assign a simple
Boolean truth value to such default rules, a semantical approach is to
define when a rational agent 
accepts such a rule. 
We could say that an agent accepts the rule \emph{Birds normally fly}
if she considers a world with a flying bird to be less surprising
than a world with a nonflying bird.
At the same time, the agent can also accept the rule
\emph{Penguin birds normally do not fly}; this is the case if she
considers a world with a nonflying penguin bird to be less surprising
than a world with a flying penguin bird.%
\fussnoteOhneMarkierung{The research reported here was partially supported by the 
Deutsche Forschungsgemeinschaft -- DFG (grants BE 1700/7-2 and KE 1413/2-2).}

The informal notions just used can be made precise by formalizing 
the underlying concepts like default rules, epistemic state of
an agent, and the acceptance relation between epistemic states and
default rules. In the following, we deal with qualitative default rules
and a corresponding semantics modelling the epistemic state of an agent.
While a full epistemic state could compare possible worlds according to
their possibility, their probability, their degree of plausibility, etc.
(cf.\ \cite{Spohn88,GoldszmidtMorrisPearl93,GoldszmidtPearl96}),
we will use ordinal conditional functions (OCFs), which are also
called ranking functions \cite{Spohn88}.
To each possible world \(\omega\), an OCF \(\kappa\) assigns
a natural number \(\kappa(\omega)\) indicating its degree of surprise:
The higher \(\kappa(\omega)\), the greater is the surprise for
observing \(\omega\).

In \cite{Kern-Isberner00d,Kern-Isberner02a} a criterion when a 
ranking function respects the conditional structure of a set \R\
of conditionals is defined, leading to the notion of
c-representation for \R, and it is argued that ranking functions
defined by c-representations
are of particular interest 
for model-based inference. 
In \cite{BeierleKernIsbernerKoch2008IJCAR} a system that
computes a c-representation for any such \R\ that is consistent
is described, but this c-representation may not be minimal.
An algorithm for computing a minimal ranking function
is given in  \cite{Bourne99PhD}, but this
algorithm fails to find all minimal ranking functions if there is more
than one minimal one.
In \cite{Mueller2004BSc} an extension of that algorithm being able
to compute all minimal 
c-representations
for \R\ is presented. The algorithm developed in \cite{Mueller2004BSc}
uses a non-declarative approach and is implemented in an imperative 
programming language.
While the problem of specifying all 
c-representations
for \R\ is formalized as an abstract,
problem-oriented
constraint satisfaction problem \cspR\ in \cite{BeierleKernIsberner2011MOC},
no solving method is given there.

In this paper, we present a high-level, declarative approach using 
constraint logic programming techniques for solving the constraint 
satisfaction problem \cspR\ for any consistent \R.
In particular, the approach developed here supports 
the generation of all minimal solutions;
these minimal solutions are of special interest as they provide a 
preferred basis for model-based inference from \R.

The rest of this paper is organized as follows:
After recalling the formal background of
conditional logics as it is given in \cite{BeierleKernIsberner2008ABZ}
and as far as it is needed here (Section~\ref{sec_background}),
we elaborate the birds-penguins scenario 
sketched above as an illustration
for a conditional knowledge base and its semantics
in Section~\ref{sec_example}.
The definition of the constraint satisfaction problem \cspR\ 
and its solution set denoting all c-representations for \R\ is
given in Sec.~\ref{sec_c_representations}.
In Section~\ref{sec_clp_approach}, a declarative, high-level CLP program 
solving \cspR\ is developed,
observing the objective of being as close as possible to \cspR, and
its realization
in 
Prolog is described in detail;
in Section~\ref{sec_Example_Applications_and_First_Evaluation}, 
it is evaluated with 
respect to a series of some first example applications.
Section~\ref{sec_conclusions} concludes the paper and points out 
further work.

\section{Background}
\label{sec_background}

\begin{sloppypar}
We start with a  propositional language $\cL$, generated by a finite set 
\Vocabulary\ of atoms $a,b,c, \ldots$. The formulas of \(\cL\) will be denoted
by uppercase Roman letters $A,B,C, \ldots$. For conciseness of notation, we
will omit the logical  
\textit{and}-connective, writing $AB$ instead of $A \wedge B$, 
and 
overlining
formulas will indicate negation, i.e.\ $\notA$ means $\neg A$. 
Let $\Omega$ denote the set of possible worlds over $\cL$; $\Omega$ will be taken here simply 
as the  set of all propositional interpretations over $\cL$ and can be
identified with the set of all complete conjunctions over \Vocabulary. 
For $\omega \in \Omega$, 
$\omega \models A$ means that  
the propositional formula $A \in \cL$ holds in the possible world 
$\omega$. 
\end{sloppypar}

By introducing a new binary operator $|$, we obtain the set 
\(
   \condL = \{ (B|A) \mid A,B \in \cL\}
\)
of \emph{conditionals} over $\cL$. $\condAB$ formalizes ``\textit{if $A$ then (normally)
$B$}'' and 
establishes a plausible,  probable, possible etc connection between the \emph{antecedent} $A$ 
and the \emph{consequence} $B$. Here, conditionals are supposed  not to be nested, that is, 
antecedent and consequent of a conditional will be propositional formulas.

A conditional \satzCL{B}{A} is an object of a three-valued nature, 
partitioning the set of worlds  $\Omega$ in three parts: 
those worlds satisfying \(AB\), thus \emph{verifying} the conditional,
those worlds satisfying \(A\ol{B}\), thus \emph{falsifying} the 
conditional, and those worlds not fulfilling the premise \(A\) and 
so which the conditional may not be applied to at all.
This allows us 
to represent \satzCL{B}{A} as a 
\emph{generalized indicator function} going back to
\cite{deFinetti74abbreviated}
(where \(u\) stands for \emph{unknown}
or \emph{indeterminate}):
\begin{equation}
\label{eq_indicator_function}
  \satzCL{B}{A}(\omega) =
\left\{
\begin{array}{l@{\; \mbox{\ \ \ \ if \ } \;}l}
1  & \omega \models AB\\
0  & \omega \models A\ol{B}\\
u  & \omega \models \ol{A}
\end{array}
\right.
\end{equation}
To give appropriate semantics to conditionals, they are usually considered
within richer structures such as \emph{epistemic states}.  
Besides certain (logical) knowledge, epistemic states also  
allow the representation of preferences, beliefs, assumptions 
of an
intelligent agent. Basically, an epistemic state allows one to compare formulas
or worlds with respect to plausibility, possibility, necessity, probability,
etc. 

Well-known qualitative, ordinal approaches to represent epistemic states
are  Spohn's \emph{ordinal conditional functions, OCFs}, (also called \emph{ranking
    functions}) \cite{Spohn88}, and \emph{possibility distributions} \cite{BenferhatDuboisPrade92},
    assigning degrees of 
    plausibility, or of possibility, respectively, to formulas and possible
    worlds. 
In such qualitative frameworks, a conditional
$(B|A)$ is valid (or \emph{accepted}), if its confirmation, $AB$, is more
plausible, possible, etc.\ than its refutation, $A\notB$; a suitable degree of
acceptance is calculated from the degrees associated with $AB$ and $A\notB$.  

In this paper, we consider
Spohn's OCFs
\cite{Spohn88}. An OCF is a function
\[
     \displayFctEins{\StateSymOCF}{\World}{\MeasureOCF}
\]
expressing degrees of plausibility of propositional formulas
where a higher degree denotes ``less plausible'' or ``more suprising''.
At least one world must be regarded as being normal; therefore,
\(
     \StateSymOCF(\omega) = 0
\)
for at least one \(\omega \in \World\).
Each such ranking function can be taken as the representation of 
a full epistemic state of an agent. 
Each such \(\StateSymOCF\)
uniquely 
extends to a function 
(also denoted by~\(\StateSymOCF\))
mapping sentences and rules to \(\MeasureOCF \cup \{\infty\}\) 
and being defined by 
%
%
\begin{equation}
\label{eq_kappa_on_formulas}
\begin{array}{lll}
  \StateSymOCF(A) & = &
    \begin{cases}     \min\{\StateSymOCF(\omega) \mid 
                      \omega \models A\} 
                          & \textrm{\ \ if } A \textrm{ is satisfiable} \cr
           \infty         & \textrm{\ \ otherwise}          
    \end{cases}
\end{array}
\\
\end{equation}
for sentences \(A \in \cL\) and by
\begin{equation}
\label{eq_kappa_on_conditionals}
\begin{array}{@{\hspace*{-2.1cm}}lll}
  \StateSymOCF(\satzCL{B}{A}) & = &
    \begin{cases} \StateSymOCF(AB) - \StateSymOCF(A)
                      & \textrm{\ \ if } \StateSymOCF(A) \not= \infty \cr
           \infty     & \textrm{\ \ otherwise}          
    \end{cases}
\end{array}
\end{equation}
for conditionals 
\(\satzCL{B}{A} \in \condL\).
Note that 
\(
   \StateSymOCF(\satzCL{B}{A}) \geq 0
\)
since any \(\omega\) satisfying \(AB\) also satisfies \(A\) and therefore
\(
   \StateSymOCF(AB) \geq \StateSymOCF(A)
\).

The belief of an agent being in epistemic state \( \StateSymOCF \)
with respect to a
default rule  \(\satzCL{B}{A}\)
is determined by the satisfaction relation 
\(
   \modelsQnoIocf
\)
defined by:
\begin{equation}
\label{equation_OCF_satisfaction_relation_unquantified}
   \StateSymOCF  \modelsQnoIocf \satzCL{B}{A}
        \textrm{ \ \ iff \ \ }
     \StateSymOCF(AB) \ < \  \StateSymOCF(A\notB)
\end{equation}
Thus, \(\satzCL{B}{A}\) is believed in \(\StateSymOCF\)
iff the rank 
of 
\(
    AB
\)
(verifying the conditional) is 
strictly smaller than the rank
of 
\(
    A\notB
\)
(falsifying the conditional).
We say that 
\(\StateSymOCF\) \emph{accepts} the conditional
\(\satzCL{B}{A}\) iff 
\( 
   \StateSymOCF  \modelsQnoIocf \satzCL{B}{A}
\).

\section{Example}
\label{sec_example}

In order to illustrate the concepts presented in the previous section 
we will use a scenario involving a set of some default rules
representing common-sense knowledge.

\begin{example}
\label{example:penguins_1}
Suppose we have the propositional atoms 
    \(f\) - \emph{flying},
    \(b\) - \emph{birds},
    \(p\) - \emph{penguins},
    \(w\) - \emph{winged} animals,
    \(k\) - \emph{kiwis}.

Let the set \({\cal R}\) consist of the following conditionals:
\\[1.2mm]
\hspace*{2.5cm}
\begin{tabular}{l@{\hspace{1cm}}l@{\ : \ \,}l@{\hspace*{0.4cm}}l}
\({\cal R}\) 
    & \(r_1\) & \satzCL{f}{b}       &  \emph{birds fly}\\
    & \(r_2\) & \satzCL{b}{p}       &  \emph{penguins are birds}\\
    & \(r_3\) & \satzCL{\ol{f}}{p}  &  \emph{penguins do not fly}\\
    & \(r_4\) & \satzCL{w}{b}       &  \emph{birds have wings}\\
    & \(r_5\) & \satzCL{b}{k}       &  \emph{kiwis are birds}
\end{tabular}
\end{example}

Figure~\ref{figure_kappa_after_initialzation}
shows a ranking function \(\kappa\) that accepts all
conditionals given in \R.
Thus, for any \(i \in \{1,2,3,4,5\}\) it holds that
\(
    \kappa \modelsQnoIocf R_i
\).

\begin{figure}[htb]
\begin{center}
\begin{tabular}{cc@{\hspace*{0.8cm}}cc@{\hspace*{0.8cm}}cc@{\hspace*{0.8cm}}cc}\toprule
\(\omega\)  &  \(\kappa(\omega)\) &
\(\omega\)  &  \(\kappa(\omega)\) &
\(\omega\)  &  \(\kappa(\omega)\) &
\(\omega\)  &  \(\kappa(\omega)\) 
\\ \midrule
\(pbfwk\)                           & 2 
   & \(p\ol{b}fwk\)                      & 5 
     &  \(\ol{p}bfwk\)                      & 0 
       &   \(\ol{p}\ol{b}fwk\)                 & 1 
\\
\(pbfw\ol{k}\)                      & 2 
   & \(p\ol{b}fw\ol{k}\)                 & 4 
      & \(\ol{p}bfw\ol{k}\)                 & 0 
         & \(\ol{p}\ol{b}fw\ol{k}\)            & 0 
\\
\(pbf\ol{w}k\)                      & 3 
   & \(p\ol{b}f\ol{w}k\)                 & 5 
      & \(\ol{p}bf\ol{w}k\)                 & 1 
         & \(\ol{p}\ol{b}f\ol{w}k\)            & 1 
\\
\(pbf\ol{w}\ol{k}\)                 & 3 
   & \(p\ol{b}f\ol{w}\ol{k}\)            & 4 
      & \(\ol{p}bf\ol{w}\ol{k}\)            & 1 
         & \(\ol{p}\ol{b}f\ol{w}\ol{k}\)       & 0 
\\[1mm]


\(pb\ol{f}wk\)                      & 1 
   & \(p\ol{b}\ol{f}wk\)                 & 3 
      & \(\ol{p}b\ol{f}wk\)                 & 1 
         & \(\ol{p}\ol{b}\ol{f}wk\)            & 1 
\\
\(pb\ol{f}w\ol{k}\)                 & 1 
   & \(p\ol{b}\ol{f}w\ol{k}\)            & 2 
      & \(\ol{p}b\ol{f}w\ol{k}\)            & 1 
         & \(\ol{p}\ol{b}\ol{f}w\ol{k}\)       & 0 
\\
\(pb\ol{f}\ol{w}k\)                 & 2 
   & \(p\ol{b}\ol{f}\ol{w}k\)            & 3 
      & \(\ol{p}b\ol{f}\ol{w}k\)            & 2 
         & \(\ol{p}\ol{b}\ol{f}\ol{w}k\)       & 1 
\\
\(pb\ol{f}\ol{w}\ol{k}\)            & 2 
   & \(p\ol{b}\ol{f}\ol{w}\ol{k}\)       & 2 
      & \(\ol{p}b\ol{f}\ol{w}\ol{k}\)       & 2 
         & \(\ol{p}\ol{b}\ol{f}\ol{w}\ol{k}\)  & 0 
\\
\bottomrule
\end{tabular}
\end{center}
\caption{Ranking function \(\kappa\) 
         accepting the rule set 
        \({\cal R}\) given in Example~\ref{example:penguins_1}}
\label{figure_kappa_after_initialzation}
\end{figure}

For the conditional \satzCL{f}{p} (\emph{``Do penguins fly?''})
that is not contained in \R,
we get
\(
   \kappa(pf) = 2
\)
and 
\(
   \kappa(p\ol{f}) = 1
\)
and therefore
\[
    \kappa \nichtMod\modelsQnoIocf \satzCL{f}{p}
\]
so that the conditional \satzCL{f}{p} is not accepted by \(\kappa\).
This is in accordance with the behaviour of a rational agent 
believing \R\ since 
the knowledge base \R\ used for building up \(\kappa\) explicitly
contains the opposite rule \(\satzCL{\ol{f}}{p}\).

On the other hand,
for the conditional \satzCL{w}{k} (\emph{``Do kiwis have wings?''})
that is also not contained in \R,
we get
\(
   \kappa(kw) = 0
\)
and 
\(
   \kappa(k\ol{w}) = 1
\)
and therefore
\[
    \kappa \modelsQnoIocf \satzCL{w}{k}
\]
i.e.,
the conditional \satzCL{w}{k} is accepted by \(\kappa\).
Thus, 
from their superclass \emph{birds},
kiwis inherit the property of having wings.

\section{Specification of Ranking Functions as Solutions of a Constraint Satisfaction Problem}
\label{sec_c_representations}

Given a set
\(
   \R = \{R_1,\ldots,R_n\}
\)
of conditionals, a ranking function \StateSymOCF\ that accepts every
\(R_i\) repesents an epistemic state of an agent accepting 
\R. 
If there is no \StateSymOCF\ that accepts every \(R_i\) then \R\
is \emph{inconsistent}. For the rest of this paper, we assume that
\R\ is consistent. 

For any consistent \R\
there may be many different \StateSymOCF\ accepting \R, each representing
a complete set of beliefs with respect to every possible formula \(A\)
and every conditional \(\satzCL{B}{A}\).
Thus, every such \StateSymOCF\
inductively completes the knowledge given by \R,
and it is a vital question whether some \(\StateSymOCF'\) is to be preferred
to some other \(\StateSymOCF''\), or whether there is a unique ``best''
\StateSymOCF.
Different ways of determining a ranking function are given by 
\emph{system Z} \cite{GoldszmidtMorrisPearl93,GoldszmidtPearl96} or its
more sophisticated extension 
\emph{system Z$^{*}$} \cite{GoldszmidtMorrisPearl93},
see also \cite{BourneParsons99};
for an approach using rational world rankings see 
\cite{Weydert98KR}.
For quantitative knowledge bases of the form
\(
   \R_x = \{\satzPCL{B_1}{A_1}{x_1},\ldots,\satzPCL{B_n}{A_n}{x_n}\}
\)
with probability values \(x_i\) and with models being probability 
distributions \(P\) satisfying a probabilistic conditional
\(
     \satzPCL{B_i}{A_i}{x_i}
\)
iff
\(
     P(B_i|A_i) = x_i
\),
a unique model can be choosen by employing the principle of maximum 
entropy \cite{Paris94,ParisVencovska97,Kern-Isberner97e};
the maximum entropy model is a best model in the sense that it is the
most unbiased one among all models satisfying \(\R_x\).

Using the maximum entropy idea, in
\cite{Kern-Isberner02a} a generalization of system Z$^{*}$ is 
suggested. Based on an algebraic treatment of conditionals, the notion
of \emph{conditional indifference} of \StateSymOCF\ with respect to \R\
is defined and the following criterion for conditional indifference
is given:
An OCF \StateSymOCF\ is  indifferent with respect to 
\(
     \R = \{\satzCL{B_1}{A_1},\ldots,\satzCL{B_n}{A_n}\}
\)
iff 
\(
     \StateSymOCF(A_i) < \infty
\)
for all \(i \in \{1,\ldots,n\}\) 
and
there are rational numbers 
\(
    \kappa_0, \kappaiplus{i}, \kappaiminus{i} \in \rationals, \ 1 \leq i \leq n,
\)
such that for all $\omega \in \Omega$,
\begin{equation}
\label{eq_conditional_indifference_ocf}
\kappa(\omega) = \kappa_0 + \sum_{1 \leq i \leq n \atop \omega \models A_i B_i}
\kappaiplus{i} + \sum_{1 \leq i \leq n \atop \omega \models A_i \ol{B_i}}
\kappaiminus{i}. 
\end{equation}
When starting with an epistemic state of complete ignorance (i.e., each world
\(\omega\) has rank 0), for each rule \(\satzCL{B_i}{A_i}\)
the values
\(
   \kappaiplus{i}, \kappaiminus{i}
\)
determine how the rank of each satisfying world and of each falsifying world, 
respectively, should be changed:
\begin{itemize}
\item
If the world \(\omega\) verifies the conditional \(\satzCL{B_i}{A_i}\),
\, -- \, i.e., 
\(
       \omega \models A_i B_i
\) 
\, --, then \(\kappaiplus{i}\) is used in the summation to obtain the value
\(\kappa(\omega)\).
\item
Likewise, if \(\omega\) falsifies the conditional \(\satzCL{B_i}{A_i}\),
\, 
-- \, i.e., 
\(
       \omega \models A_i \ol{B_i}
\) 
\, --, then \(\kappaiminus{i}\) is used in the summation instead.
\item
If the conditional \(\satzCL{B_i}{A_i}\) is not applicable
in \(\omega\),
\, -- \, i.e., 
\(
       \omega \models \ol{A_i} 
\)
\, --, then this conditional does not influence 
the value \(\kappa(\omega)\).
\end{itemize}

\(\kappa_0\) is a normalization
constant ensuring that there is a smallest world rank 0.
Employing the postulate that the ranks of a satisfying world 
should not be changed
and requiring that changing the rank of a falsifying world
may not result in an increase of the world's plausibility 
leads to the concept of a \emph{c-representation}
\cite{Kern-Isberner02a,Kern-Isberner00d}:
\begin{definition}
\label{def_c_representation}
Let
\(
     \R = \{\satzCL{B_1}{A_1},\ldots,\satzCL{B_n}{A_n}\}
\).
Any ranking function \StateSymOCF\ satisfying
the conditional indifference condition
(\ref{eq_conditional_indifference_ocf})
and
\(
      \kappaiplus{i} = 0
\),
\(
     \kappaiminus{i} \geq 0
\)
(and thus also
\(
   \kappa_0 = 0
\) since \R\ is assumed to be consistent)
as well as
\begin{equation}
\label{eq_kappa_accepts_r}
     \StateSymOCF(A_iB_i)   <   \StateSymOCF(A_i\ol{B_i})
\end{equation}
for all \(i \in \{1,\ldots,n\}\)
is called a \emph{(special) c-representation} of \R.
\end{definition}
Note that for \(i \in \{1,\ldots,n\}\),
condition (\ref{eq_kappa_accepts_r}) expresses that \(\kappa\)
accepts the conditional \(R_i = \satzCL{B_i}{A_i} \in \R\)
(cf.\ the definition of the satisfaction relation in 
(\ref{equation_OCF_satisfaction_relation_unquantified}))
and that this 
also
implies \(\StateSymOCF(A_i) < \infty\).

Thus, finding a c-representation for \(\R\) amounts to choosing appropriate
values \( \kappaiminus{1}\), \ldots,  \( \kappaiminus{n} \).
In \cite{BeierleKernIsberner2011MOC} this situation is formulated as a 
constraint satisfaction problem \cspR\ whose solutions are 
vectors of the form \( (\kappaiminus{1}, \ldots,  \kappaiminus{n}) \)
determining
c-representations of \R.
The development of \cspR\ exploits 
 (\ref{eq_kappa_on_formulas})
and
 (\ref{eq_conditional_indifference_ocf})
to reformulate
 (\ref{eq_kappa_accepts_r})
and
requires that the 
\(\kappaiminus{i}\)
are natural numbers (and not just rational numbers).
In the following, we set \(\min(\emptyset) = \infty\).
%
\begin{definition}\textbf{[\boldmath{\cspR}]} \ 
\label{def_csp_fuer_r}
Let
\(
     \R = \{\satzCL{B_1}{A_1},\ldots,\satzCL{B_n}{A_n}\}
\).
The constraint satisfaction problem for c-representations of \R,
denoted by \cspR, is given by the conjunction of the constraints
%
\begin{align}
\label{eq_kappaiminus_positive}
&
\kappaiminus{i} \geq 0
\\
%
\label{eq_kappa_accepts_r_with_kappaiminus}
&
\kappaiminus{i}  >  
   \min_{\omega \models A_i B_i}
           \sum_{j \neq i \atop \omega \models A_j \ol{B_j}} \kappaiminus{j} 
    - 
   \min_{\omega \models A_i \notB_i}
           \sum_{j \neq i \atop \omega \models A_j \ol{B_j}} \kappaiminus{j} 
\end{align}
for all \(i \in \{1,\ldots,n\}\).
\end{definition}
A solution of \cspR\ is an \(n\)-tupel
\(
     (\kappaiminus{1}, \ldots,  \kappaiminus{n})
\)
of natural numbers, and with
\solutionsR\ we denote the set of all solutions of \cspR.
\begin{proposition}
\label{prop_solutions_csp}
For 
\(
     \R = \{\satzCL{B_1}{A_1},\ldots,\satzCL{B_n}{A_n}\}
\)
let 
\(
     (\kappaiminus{1}, \ldots,  \kappaiminus{n})  \in  \solutionsR
\).
Then the function \(\kappa\) defined by 
\begin{equation}
\label{eq_conditional_indifference_no_kappaplus}
\kappa(\omega) = 
\sum_{1 \leq i \leq n \atop \omega \models A_i \ol{B_i}}
\kappaiminus{i} 
\end{equation}
accepts \R.
\end{proposition}

All c-representations built from 
(\ref{eq_kappaiminus_positive}),
(\ref{eq_kappa_accepts_r_with_kappaiminus}), and
(\ref{eq_conditional_indifference_no_kappaplus})
provide an excellent basis for model-based inference 
\cite{Kern-Isberner02a,Kern-Isberner00d}.
However, from the point of view of minimal specificity
(see e.g.\ \cite{BenferhatDuboisPrade92}), those  c-representations
with minimal \(\kappaiminus{i}\) yielding minimal degrees of
implausibility are most interesting.

While different orderings on \solutionsR\ can be defined,
leading to different minimality notions, 
in the following we will focus on the ordering on \solutionsR\
induced by taking the sum of the \(\kappaiminus{i}\), i.e.\
\begin{equation}
\label{eq_ordering_kappa_vectors_sum}
    (\kappaiminus{1}, \ldots,  \kappaiminus{n})
                     \leq
    (\kappaiminusStrich{1}, \ldots,  \kappaiminusStrich{n})
             \quad \textrm { iff }  \quad
    \sum_{1 \leq i \leq n} \kappaiminus{i}
                     \leq
    \sum_{1 \leq i \leq n} \kappaiminusStrich{i}.
\end{equation}

As we are interested in minimal \(\kappaiminus{i}\)-vectors, an important
question is whether there is always a unique minimal solution.
This is not the case; 
the following example that is also discussed in
\cite{Mueller2004BSc} illustrates that 
\solutionsR\ may have more than one minimal element.
\begin{example}
\label{eq_multiple_minimal_solutions}
Let \(\R_{\mathit{birds}} = \{R_1, \, R_2, \, R_3\}\) be the following set of conditionals:
\[
\begin{array}{ll@{\hspace{1.5cm}}l}
R_1: & \satzCL{f}{b} & \textrm{\emph{\underline{b}irds \underline{f}ly}}
\\
R_2: & \satzCL{a}{b} & \textrm{\emph{\underline{b}irds are \underline{a}nimals}}
\\
R_3: & \satzCL{a}{fb} & \textrm{\emph{\underline{f}lying \underline{b}irds are \underline{a}nimals}}
\end{array}
\]
From (\ref{eq_kappa_accepts_r_with_kappaiminus}) we get
\[
\begin{array}{ll@{\hspace{1.5cm}}l}
\kappaiminus{1} > 0
\\
\kappaiminus{2} > 0 - min\{\kappaiminus{1}, \, \kappaiminus{3}\}
\\
\kappaiminus{3} > 0 - \kappaiminus{2}
\end{array}
\]
and since \(\kappaiminus{i} \geq 0\) according to 
(\ref{eq_kappaiminus_positive}), the two vectors
\[
\begin{array}{ll@{\hspace{2cm}}l}
\mathit{sol}_1 = (\kappaiminus{1}, \kappaiminus{2}, \kappaiminus{3})
               = (1, 1, 0)
\\
\mathit{sol}_2 = (\kappaiminus{1}, \kappaiminus{2}, \kappaiminus{3})
               = (1, 0, 1)
\end{array}
\]
are two different 
solutions of 
\(\cspOf{\R_{\mathit{birds}}}\) 
with
\(
    \sum_{1 \leq i \leq n} \kappaiminus{i} = 2
\)
that are both minimal in 
\(\solutionsOf{\R_{\mathit{birds}}}\) with respect to 
\(\leq\).
\end{example}

\section{A Declarative CLP Program for \cspR}
\label{sec_clp_approach}

In this section, we will develop a CLP program \cspRprog\
solving \cspR.
Our main objective to obtain a declarative program that is
as close as possible to the abstract formulation of \cspR\
while exploiting the concepts of constraint logic programming.
We will employ  finite domain constraints, and from 
(\ref{eq_kappaiminus_positive}) we immediately get a lower bound
for \(\kappaiminus{i}\). 
Considering that we are interested mainly in minimal solutions,
due to (\ref{eq_kappa_accepts_r_with_kappaiminus}) we can safely restrict
ourselves to \(n\) as an upper bound for  \(\kappaiminus{i}\),
yielding
\begin{align}
\label{eq_kappaiminus_lower_upper_bound}
     0 \leq \kappaiminus{i} \leq n
\end{align}
for all \(i \in \{1,\ldots,n\}\) 
with \(n\) being the number of conditionals in \R.

\subsection{Input Format and Preliminaries}

Since we want to focus on the constraint solving part, we do not
consider reading and parsing a knowledge base 
\(
     \R = \{\satzCL{B_1}{A_1},\ldots,\satzCL{B_n}{A_n}\}
\).
Instead, we assume
that 
\R\
is already given as a Prolog code file providing the following
predicates 
\textaa{variables/1},
\textaa{conditional/3} and \textaa{indices/1}:
\\[\abstand]
\hspace*{0.5cm}
\begin{tabular}{l@{\hspace{0.4cm}}l}
\textaa{variables([\(a_1\),\ldots,\(a_m\)])} 
    & \verb|%| list of atoms in \(\Sigma\)
\\
\textaa{conditional(\(i\),\textmeta{A_i},\textmeta{B_i})} 
    & \verb|%| representation of \(i\)th conditional  \(\satzCL{B_i}{A_i}\)
\\
\textaa{indices([1,...,\(n\)])}
    & \verb|%| list of indices \(\{1,\ldots,n\}\)
\end{tabular}
\\[\abstand]
If \(\Sigma = \{a_1,\ldots, a_m\}\) is the set of atoms, we assume
a fixed ordering 
\(
   a_1 < a_2 < \ldots < a_m
\)
on \(\Sigma\) given by the predicate
\textaa{variables([\(a_1\),\ldots,\(a_m\)])}.

In the representation of a conditional, a propositional formula
\(A\), constituting the antecedent or the consequence of the 
conditional, is represented by \textmeta{A}
where \textmeta{A} is a Prolog list
    \textaa{[\textmeta{D_1},\ldots,\textmeta{D_l}]}.
Each \textmeta{D_i} represents a conjunction of literals 
such that 
\(
    D_1 \vee \ldots \vee D_l
\)
is a disjunctive normal form of \(A\).

Each \textmeta{D}, representing a conjunction of literals, is a Prolog list
    \textaa{[\(b_1\),\ldots,\(b_m\)]}
of fixed length \(m\) where \(m\) is the number of atoms in \(\Sigma\)
and with \(b_k \in \{\textaa{0, 1, \url{\_}}\}\).
Such a list 
    \textaa{[\(b_1\),\ldots,\(b_m\)]}
represents the conjunctions of atoms obtained from
\(
       \dot{a}_1 \wedge \dot{a}_2 \wedge \ldots \wedge \dot{a}_m
\)
by eliminating all occurrences of \(\top\),
where
\[
\begin{array}{lll}
  \dot{a}_k & = &
    \begin{cases}     
           a_k            & \textrm{\ \ if } b_k = 1 \cr
           \ol{a_k}       & \textrm{\ \ if } b_k = 0 \cr
           \top           & \textrm{\ \ if } b_k = \_ 
    \end{cases}
\end{array}   
\]

\begin{example}
The internal representation of the knowledge base presented in
Example~\ref{example:penguins_1} is shown in 
Figure~\ref{fig_penguins_kiwis_internal}.
\end{example}

\begin{figure}[htb]
\begin{center}
\begin{qverbatim}
variables([p,b,f,w,k]).

conditional(1,[[_,1,_,_,_]],[[_,_,1,_,_]]).   
conditional(2,[[1,_,_,_,_]],[[_,1,_,_,_]]).   
conditional(3,[[1,_,_,_,_]],[[_,_,0,_,_]]).   
conditional(4,[[_,1,_,_,_]],[[_,_,_,1,_]]).   
conditional(5,[[_,_,_,_,1]],[[_,1,_,_,_]]).   

indices([1,2,3,4,5]).
\end{qverbatim}
\end{center}
\caption{Internal representation of the knowledge base from Example~\ref{example:penguins_1}}
\label{fig_penguins_kiwis_internal}
\end{figure}

As further preliminaries, 
using \textaa{conditional/3} and \textaa{indices/1},
we have implemented the predicates
\textaa{verifying\_worlds/2},
\textaa{falsifying\_worlds/2},
and
\textaa{falsify/2}, 
realising the evaluation of the indicator function (\ref{eq_indicator_function})
given in Sec.~\ref{sec_background}:
\\[\abstand]
\hspace*{0.5cm}
\begin{tabular}{l@{\hspace{0.3cm}}l}
\textaa{verifying\_worlds(\(i\),\(\mathit{Ws}\))} 
    & \verb|%| \(\mathit{Ws}\) list of worlds verifying \(i\)th conditional
\\
\textaa{falsifying\_worlds(\(i\),\(\mathit{Ws}\))} 
    & \verb|%| \(\mathit{Ws}\) list of worlds falsifying \(i\)th conditional
\\
\textaa{falsify(\(i\),\(W\))} 
    & \verb|%| world \(W\) falsifies \(i\)th conditional
\end{tabular}
\\[\abstand]
where worlds are represented as complete conjunctions of literals over 
\(\Sigma\), using the representation described above.

Using these predicates, in the following subsections
we will present the complete source code of the constraint logic program 
\cspRprog\
solving \cspR.

\subsection{Generation of Constraints}

The particular program code given here uses the SICStus Prolog system%
\footnote{\url{http://www.sics.se/isl/sicstuswww/site/index.html}}
 and its
clp(fd) library implementing constraint logic programming over finite domains
\cite{CarlssonOttossonCarlson97sicstusCLPfd}.

The main predicate \textaa{kappa/2} expecting a knowledge base \textaa{KB}
of conditionals and yielding a vector \textaa{K} of \(\kappaiminus{i}\)
values as specified by 
(\ref{eq_kappa_accepts_r_with_kappaiminus})
is presented in Fig.~\ref{fig_main_kappa}.
\begin{figure}[htb]
\begin{center}
\begin{qverbatim}
kappa(KB, K) :-         
   consult(KB),
   indices(Is),         
   length(Is, N),       
   length(K, N),        
   domain(K, 0, N),     
   constrain_K(Is, K),  
   labeling([], K).     
\end{qverbatim}
\end{center}
\caption{Main predicate \textaa{kappa/2}}
\label{fig_main_kappa}
\end{figure}

After reading in the knowledge base and getting the list of indices, a list 
\textaa{K} of free constraint variables, one for each conditional, is
generated.
In the two subsequent subgoals, the constraints corresponding to
the formulas 
(\ref{eq_kappaiminus_lower_upper_bound})
and
(\ref{eq_kappa_accepts_r_with_kappaiminus})
are generated, constraining the elements of \textaa{K} accordingly.
Finally, \textaa{labeling([], K)} yields a list of  \(\kappaiminus{i}\)
values. Upon backtracking, this will enumerate all possible solutions
with an upper bound of \(n\) as in 
(\ref{eq_kappaiminus_lower_upper_bound})
for each \(\kappaiminus{i}\).
Later on, we will demonstrate how to modify \textaa{kappa/2}
in order to take minimality into account 
(Sec.~\ref{sec_Generation_of_Minimal_Solutions}).

How the
subgoal \textaa{constrain\_K(Is, K)}
in \textaa{kappa/2} generates a constraint for each index
\(i \in \{1,\ldots,n\}\) according to 
(\ref{eq_kappa_accepts_r_with_kappaiminus})
is defined in Fig.~\ref{fig_constrain_k}.

\begin{figure}[htb]
\begin{qverbatim}
constrain_K([],_).                      
constrain_K([I|Is],K) :-                
   constrain_Ki(I,K), constrain_K(Is,K).

constrain_Ki(I,K) :-           
   verifying_worlds(I, VWorlds),     
   falsifying_worlds(I, FWorlds),    
   list_of_sums(I, K, VWorlds, VS),  
   list_of_sums(I, K, FWorlds, FS),  
   minimum(Vmin, VS),                
   minimum(Fmin, FS),                
   element(I, K, Ki),                
   Ki #> Vmin - Fmin.                
\end{qverbatim}
\caption{Constraining the vector \textaa{K} representing \(\kappaiminus{1},\ldots,\kappaiminus{n}\) as in (\ref{eq_kappa_accepts_r_with_kappaiminus})}
\label{fig_constrain_k}
\end{figure}

Given an index \textaa{I}, \textaa{constrain\_Ki(I,K)}
determines all worlds verifying and falsifying the \textaa{I}-th
conditional; over these two sets of worlds the two \(\min\) expressions
in
(\ref{eq_kappa_accepts_r_with_kappaiminus})
are defined.
Two lists \textaa{VS} and \textaa{FS}  
of sums corresponding exactly to the first and the second sum,
repectively, in 
(\ref{eq_kappa_accepts_r_with_kappaiminus}) 
are generated (how this is done is defined in
Fig.~\ref{fig_list_of_sums} and will be explained below). 
With the constraint variables \textaa{Vmin} and \textaa{Fmin}
denoting the minimum of these two lists, the constraint
\begin{center}
     \verb|Ki #> Vmin - Fmin|         
\end{center}
given in the last line of Fig.~\ref{fig_constrain_k}
reflects precisely the restriction on   \(\kappaiminus{i}\)
given by
(\ref{eq_kappa_accepts_r_with_kappaiminus}).

For an index \textaa{I}, a kappa vector \textaa{K}, and a list of
worlds \textaa{Ws}, the 
goal
     \textaa{list\_of\_sums(I, K, Ws, Ss)}
(cf.~Fig.~\ref{fig_list_of_sums})
yields a list \textaa{Ss} of sums such that for each world 
\textaa{W} in \textaa{Ws}, there is a sum \textaa{S} in \textaa{Ss}
that is generated by 
   \textaa{sum\_kappa\_j(Js, I, K, W, S)}
where \textaa{Js} is the list of indices 
\(\{1,\ldots,n\}\).
In the goal
      \textaa{sum\_kappa\_j(Js, I, K, W, S)},
\textaa{S} corresponds exactly to the respective sum expression in 
(\ref{eq_kappa_accepts_r_with_kappaiminus}),
i.e., it is the sum of all \textaa{Kj} such that 
      \textaa{J \(\neq\) I}
and \textaa{W} falsifies the \textaa{j}-th conditional.
\begin{figure}
\begin{qverbatim}

list_of_sums(_, _, [], []).
list_of_sums(I, K, [W|Ws], [S|Ss]) :- 
   indices(Js),
   sum_kappa_j(Js, I, K, W, S), 
   list_of_sums(I, K, Ws, Ss).


sum_kappa_j([], _, _, _, 0).
sum_kappa_j([J|Js], I, K, W, S) :- 
    sum_kappa_j(Js, I, K, W, S1),
    element(J, K, Kj),
    ((J \= I, falsify(J, W)) -> S #= S1 + Kj; S #= S1).    
\end{qverbatim}
\caption{Generating list of sums of \(\kappaiminus{i}\) as in (\ref{eq_kappa_accepts_r_with_kappaiminus})}
\label{fig_list_of_sums}
\end{figure}

\begin{example}
\label{ex_kb_birds_clp_output}
Suppose that \textaa{kb\_birds.pl} is a file containing 
the conditionals of the knowledge base \(\R_{\mathit{birds}}\)
given in Ex.~\ref{eq_multiple_minimal_solutions}.
Then the first five solutions generated by the program 
given in Figures \ref{fig_main_kappa} -- \ref{fig_list_of_sums}
are:
\begin{verbatim}
     | ?- kappa('kb_birds.pl', K).
     K = [1,0,1] ? ;
     K = [1,0,2] ? ;
     K = [1,0,3] ? ;
     K = [1,1,0] ? ;
     K = [1,1,1] ? 
\end{verbatim}
Note that the first and the fourth solution are the minimal
solutions.
\end{example}

\begin{example}
\label{example:penguins_1_clp_output}
If \textaa{kb\_penguins.pl} is a file containing 
the conditionals of the knowledge base \(\R\)
given in Ex.~\ref{example:penguins_1},
the first six solutions generated by 
\texttt{kappa/2}
are:
\begin{verbatim}
     | ?- kappa('kb_penguins.pl', K).
     K = [1,2,2,1,1] ? ;
     K = [1,2,2,1,2] ? ;
     K = [1,2,2,1,3] ? ;
     K = [1,2,2,1,4] ? ;
     K = [1,2,2,1,5] ? ;
     K = [1,2,2,2,1] ? 
\end{verbatim}
\end{example}

\subsection{Generation of Minimal Solutions}
\label{sec_Generation_of_Minimal_Solutions}

The enumeration predicate \textaa{labeling/2} of SICStus Prolog allows
for an option that minimizes the value of a cost variable.
Since we are aiming at minimizing the sum of all \(\kappaiminus{i}\),
the constraint \textaa{sum(K, \#=, S)} introduces such a cost variable
\textaa{S}. Thus, exploiting the SICStus Prolog minimization feature,
we can easily modify \textaa{kappa/2} to generate a minimal solution:
We just have to replace the last subgoal \textaa{labeling([], K)}
in Fig.~\ref{fig_main_kappa} by the two subgoals:
\begin{verbatim}
   sum(K, #=, S),               % introduce constraint variable S 
                                %    for sum of kappa_I
   minimize(labeling([],K), S). % generate single minimal solution
\end{verbatim}
With this modification, we obtain a predicate \textaa{kappa\_min/2} that
returns a single minimal solution (and fails on backtracking).
Hence calling 
   \textaa{?- kappa\_min('kb\_birds.pl', K).}
similar as in Ex.~\ref{ex_kb_birds_clp_output} yields the minimal
solution
      \textaa{K = [1,0,1]}.

However, as pointed out in Sec.~\ref{sec_c_representations}, there are
good reasons for considering not just a single minimal solution, but
all minimal solutions. We can achieve the computation of all minimal
solutions by another slight modification of \textaa{kappa/2}.
This time, the enumeration subgoal \textaa{labeling([], K)}
in Fig.~\ref{fig_main_kappa} is 
preceded by two new subgoals  
as in  \textaa{kappa\_min\_all/2} in Fig.~\ref{fig_main_kappa_min_all}. 
\begin{figure}[htb]
\begin{qverbatim}
kappa_min_all(KB, K) :-  
   consult(KB),
   indices(Is),          
   length(Is, N),        
   length(K, N),         
   domain(K, 0, N),      
   constrain_K(Is, K),   
   sum(K, #=, S),        
   min_sum_kappas(K, S), 
   labeling([], K).      

min_sum_kappas(K, Min) :-      
   once((labeling([up],[Min]), 
         \+ \+ labeling([],K))).
\end{qverbatim}
\caption{Predicate \textaa{kappa\_min\_all/2} generating exactly all minimal solutions}
\label{fig_main_kappa_min_all}
\end{figure}

The first new subgoal 
   \textaa{sum(K, \#=, S)} 
introduces a constraint variable \textaa{S} just as in \textaa{kappa\_min/2}. 
In the subgoal 
   \textaa{min\_sum\_kappas(K, S)},
this variable \textaa{S} is constrained to the sum of a minimal solution
as determined by 
  \textaa{min\_sum\_kappas(K, Min)}.
These two new subgoals ensure that in the generation caused by the final 
subgoal \textaa{labeling([], K)}, exactly all minimal solutions are
enumerated.

\begin{example}
\label{ex_kb_birds_clp_output_all_min}
Continuing Example~\ref{ex_kb_birds_clp_output}, calling
\begin{verbatim}
     | ?- kappa_min_all('kb_birds.pl', K).
     K = [1,0,1] ? ;
     K = [1,1,0] ? ;
     no
\end{verbatim}
yields the two minimal solutions for \(\R_{\mathit{birds}}\).
\end{example}

\begin{example}
\label{example:penguins_1_clp_output_all_min}
For the situation in Ex.~\ref{example:penguins_1_clp_output},
\texttt{kappa\_min\_all/2} reveals that there is a unique minimal
solution:
\begin{verbatim}
     | ?- kappa_min_all('kb_penguins.pl', K).
     K = [1,2,2,1,1] ? ;
     no
\end{verbatim}
Determining the OCF \(\kappa\) induced by the vector
\(
(\kappaiminus{1}, \kappaiminus{2}, \kappaiminus{3},
                  \kappaiminus{4}, \kappaiminus{5})
     =  (1, 2, 2, 1, 1)
\)
according to (\ref{eq_conditional_indifference_no_kappaplus})
yields the ranking function given in 
Fig.~\ref{figure_kappa_after_initialzation}.
\end{example}

\section{Example Applications and First Evaluation}
\label{sec_Example_Applications_and_First_Evaluation}

Although the objective in developing \cspRprog\ was on being
as close as possible to the abstract formulation of the
constraint satisfaction problem \cspR, we will present the results
of some first example applications we have carried out.

For \(n \geq 1\), we generated synthetic knowledge bases
\hbox{\syntheticKB{n}} according 
to the following schema: 
Using the variables 
\(
   \{f\} \cup \{a_1, \ldots, a_n\}
\),
\syntheticKB{n} contains the \(2*n -1\)
conditionals given by::
\[
\begin{array}{lll}
\satzCL{f}{a_i}
                     & \textrm{\ \ if } i \textrm{ is odd,\ } 
                               i \in \{1,\ldots,n\}  
\\
\satzCL{\ol{f}}{a_i}
                     & \textrm{\ \ if } i \textrm{ is even,\ } 
                               i \in \{1,\ldots,n\}        
\\
\satzCL{a_i}{a_{i+1}} & \textrm{\ \ if } i \in \{1,\ldots,n-1\} 
\end{array}
\]
For instance, \syntheticKBzahl{4}{7} uses the five variables
\(\{f, a_1, a_2, a_3, a_4\}\) and contains the seven conditionals:
%
%
\[
\begin{array}{l}
\satzCL{f}{a_1} 
\\
\satzCL{\ol{f}}{a_2}
\\                  
\satzCL{f}{a_3}
\\
\satzCL{\ol{f}}{a_4}
\\
\satzCL{a_1}{a_2}
\\
\satzCL{a_2}{a_3}
\\         
\satzCL{a_3}{a_4}
\\
\end{array}
\]
The basic idea underlying the construction of these synthetic knowledge
bases \hbox{\syntheticKB{n}} is to establish a kind of subclass relationship
between \(a_{i+1}\) and  \(a_i\) for each 
\(i \in \{1,\ldots,n-1\}\) on the one hand, and to state that every
\(a_{i+1}\) is exceptional to \(a_i\) with respect to its behaviour regarding
\(f\), again for each \(i \in \{1,\ldots,n-1\}\).
This sequence of pairwise exceptional elements will force any minimal
solution of \cspOf{\syntheticKB{n}}
to have at least one \(\kappaiminus{i}\) value of size greater or 
equal to \(n\).

From \syntheticKBsymbol{n}{m}, the knowledge bases 
\syntheticKBsymbol{n}{m\!\!-\!\!j} 
are generated for \(j \in \{1,\ldots, m-1\}\) by removing the
last \(j\) conditionals.
For instance, \syntheticKBzahl{4}{5} is obtained from
\syntheticKBzahl{4}{7} by removing the two conditionals
\(
\{
\satzCL{a_2}{a_3}
\),
\(
\satzCL{a_3}{a_4}
\}
\).

\begin{figure}
\centerline{\includegraphics[width=10.72cm]{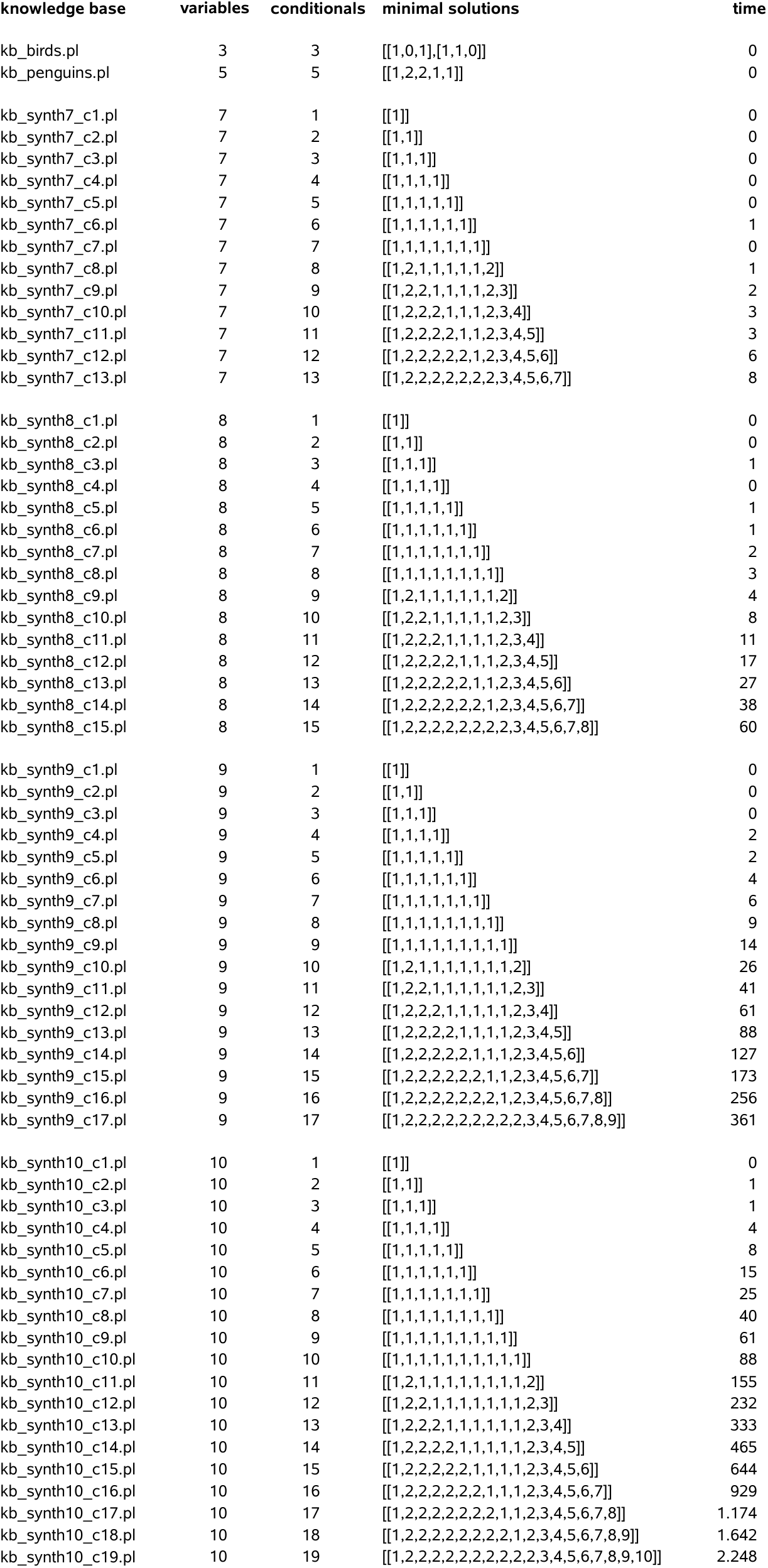}}

\caption{Execution times of \cspRprog\ under SICStus Prolog for various knowledge bases}
\label{fig_sicstus_results}
\end{figure}

Figure~\ref{fig_sicstus_results} shows the time needed by \cspRprog\ 
for computing all minimal solutions for various knowledge bases.
The execution time is given in seconds where the value 0 stands for 
any value less than 0.5 seconds.
Measurements were taken for the following environment:
SICStus 4.0.8 (x86-linux-glibc2.3), 
\hbox{Intel} Core~2~Duo E6850 3.00GHz.
While the number of variables determines the set of possible worlds,
the number of conditionals induces the number of contraints.
The values in the
table in Fig.~\ref{fig_sicstus_results}
give some indication on the influence of 
both values, the number of variables and 
the number of conditionals in a knowledge base.
For instance, 
comparing the knowledge base \syntheticKBzahl{7}{10}, 
having 8 variables and 10 conditionals, 
to  the knowledge base \syntheticKBzahl{8}{10}, 
having 9 variables and also 10 conditionals,
we see an increase of the computation time by a factor 2.3.
Increasing the number of conditionals, leads to 
no time increase from \syntheticKBzahl{7}{10} to  \syntheticKBzahl{7}{11},
and to a time increase factor
of about 1.6 when moving from \syntheticKBzahl{8}{10} to
\syntheticKBzahl{8}{11},
while for moving from
\syntheticKBzahl{8}{10} to 
\syntheticKBzahl{9}{10} and \syntheticKBzahl{10}{10},
we get time increase factors of 3.3 and 11.0, respectively.

Of course, these knowledge bases are by no means representative,
and further evaluation is needed.
In particular, investigating the complexity depending on
the number of variables and conditionals and determining an upper bound
for worst-case complexity has still to be done.
Furthermore,
while the code for \cspRprog\ given above
uses SICStus Prolog, we also have a variant of \cspRprog\
for the SWI Prolog system%
\footnote{\url{http://www.swi-prolog.org/index.html}}
\cite{WielemakerSchrijversTriskaLager2010}. 
In our further investigations, we want to evaluate \cspRprog\
also using  SWI Prolog, to elaborate the
changes required and the options provided when moving between 
SICStus and SWI Prolog, and to study  
whether there are any significant differences in execution
that might depend on the two different Prolog systems and their
options.

\section{Conclusions and Further Work}
\label{sec_conclusions}

While for a set of probabilistic conditionals
   \(\satzPCL{B_i}{A_i}{x_i}\)
the principle of maximum entropy yields a unique model, for
a set \R\ of qualitative default rules
   \(\satzCL{B_i}{A_i}\)
there may be several minimal ranking functions.
In this paper, we developed a CLP approach for solving  \cspR,
realized in the Prolog program \cspRprog. The solutions
of the constraint satisfaction problem \cspR\ are vectors of
natural numbers
\(
    \kappaiminusVektor =  (\kappaiminus{1}, \ldots,  \kappaiminus{n})
\)
that uniquely determine an OCF 
\(
       \induzierteOCF{\kappaiminusVektor}
\)
accepting all conditionals in \R.
The program
\cspRprog\ is also able to generate exactly all minimal solutions of
\cspR; the minimal solutions of \cspR\ are of special interest for
model-based inference.

Among the extentions of the approach described here we are currently
working on, is the investigation and evaluation of alternative
minimality criteria. Instead of ordering the vectors
\(
       \kappaiminusVektor
\)
by the sum of their components, we could define a componentwise
order on \solutionsR\ by defining
\(
            (\kappaiminus{1}, \ldots,  \kappaiminus{n})
                    \preceq
            (\kappaiminusStrich{1}, \ldots,  \kappaiminusStrich{n})
\)
iff \(\kappaiminus{i} \leq \kappaiminusStrich{i}\) for \(i \in \{1,\ldots,n\}\),
yielding a partial order \(\preceq\) on \solutionsR.

Still another alternative is to compare the full OCFs
\(
       \induzierteOCF{\kappaiminusVektor}
\)
induced by
\(
    \kappaiminusVektor =  (\kappaiminus{1}, \ldots,  \kappaiminus{n})
\)
according to (\ref{eq_conditional_indifference_no_kappaplus}),
yielding the ordering
\(
     \ordnungInduzierteOCF
\)
on \solutionsR\ defined by
\(
     \induzierteOCF{\kappaiminusVektor}  \ordnungInduzierteOCF 
                \induzierteOCF{\kappaiminusVektorStrich}
\)
iff
\(
      \induzierteOCF{\kappaiminusVektor}(\omega)
            \leq
      \induzierteOCF{\kappaiminusVektorStrich}(\omega)
\)
for all \(\omega \in \Omega\).

In general, it is an open problem how to strengthen
the requirements defining a c-representation so that a unique
solution is guaranteed to exist. 
The declarative nature of constraint logic programming supports
easy constraint modification, enabling the experimentation and
practical evaluation of different notions of minimality for
\solutionsR\ and of
additional requirements that might be imposed on a ranking function.
Furthermore, in \cite{EiterLukasiewicz2000AI} the framework of
default rules concidered here is extended by allowing not only
default rules in the knowledge base \R, but also strict knowledge,
rendering some worlds completely impossibe. This can yield a reduction 
of the problem's complexity, and it will be interesting to see 
which effects the incorporation of strict knowledge will have on
the CLP approach presented here.  

\bibliographystyle{plain}      

\end{document}